\documentclass[journal]{IEEEtran}
\ifCLASSOPTIONcompsoc
  \usepackage[nocompress]{cite}
\else
  \usepackage{cite}
\fi
\ifCLASSINFOpdf
\else
\fi
\usepackage{booktabs}
\usepackage{graphicx}
\usepackage{enumerate}
\usepackage{subfigure}
\usepackage{cite}

\usepackage{algorithm}
\usepackage{algorithmic}
\usepackage{graphics}

\usepackage{indentfirst}
\usepackage{verbatim}

\usepackage{amssymb}
\usepackage{longtable}
\usepackage{epsfig} 
\usepackage{amsthm}
\usepackage{mathrsfs}
\usepackage{amsmath}
\usepackage{stfloats}
\usepackage{multirow}
\usepackage{caption}
\usepackage[hang,flushmargin]{footmisc}

\begin{document}
%
\title{Towards Explainable Multi-Party Learning: A Contrastive Knowledge Sharing Framework}

%
%
%

\author{Yuan~Gao,
  Jiawei~Li,
  Maoguo~Gong,~\IEEEmembership{Senior Member,~IEEE,}
  Yu~Xie,
  and~A.~K.~Qin,~\IEEEmembership{Senior Member,~IEEE}
\thanks{Yuan Gao, Jiawei Li and Maoguo Gong are with the School of Electronic Engineering, Key Laboratory of Intelligent Perception and Image Understanding of Ministry of Education, Xidian University, Xi'an, Shaanxi Province 710071, China. (e-mail: cn\_gaoyuan@foxmail.com; Jisoo\_Lee@163.com; gong@ieee.org)}
\thanks{Yu Xie is with the Key Laboratory of Computational Intelligence and Chinese Information Processing of Ministry of Education, Shanxi University, Taiyuan 030006, China. (e-mail: sxlljcxy@gmail.com)}
\thanks{A. K. Qin is with the department of Computer Science and Software Engineering, Swinburne University of Technology, Melbourne, Australia. (e-mail: kqin@swin.edu.au)}
}

%
%

\markboth{}%
{Shell \MakeLowercase{\textit{et al.}}: Bare Demo of IEEEtran.cls for Computer Society Journals}
%



\maketitle%
\begin{abstract}
Multi-party learning provides solutions for training joint models with decentralized data under legal and practical constraints. However, traditional multi-party learning approaches are confronted with obstacles such as system heterogeneity, statistical heterogeneity, and incentive design. How to deal with these challenges and further improve the efficiency and performance of multi-party learning has become an urgent problem to be solved. In this paper, we propose a novel contrastive multi-party learning framework for knowledge refinement and sharing with an accountable incentive mechanism. Since the existing parameter averaging method is contradictory to the learning paradigm of neural networks, we simulate the process of human cognition and communication, and analogize multi-party learning as a many-to-one knowledge sharing problem. The approach is capable of integrating the acquired explicit knowledge of each client in a transparent manner without privacy disclosure, and it reduces the dependence on data distribution and communication environments. The proposed scheme achieves significant improvement in model performance in a variety of scenarios, as we demonstrated through experiments on several real-world datasets. 

\end{abstract}




%
\IEEEpeerreviewmaketitle

\section{Introduction}\label{para:1}

%
%
%
%

\IEEEPARstart{T}{he} waves of advances in the applications of deep learning keep on coming, and the foundation of these artificial intelligence algorithms is a large amount of data. However, with the enhancement of people's awareness of privacy and the improvement of relevant laws and regulations, data-driven artificial intelligence faces with serious obstacles, i.e., the limited and scattered data silos. Therefore, it has been increasingly challenging to build efficient joint models while meeting privacy and regulatory requirements. To handle this problem, a novel learning paradigm among multiple parties \cite{Bonawitz2019TowardsFL} \cite{Yang2019FederatedML} is proposed, which enables participants to train a global model with non-shared data through updating local model parameters.

There has been a growing interest in multi-party learning frameworks \cite{Yang2020FederatedLV} \cite{Gong2020PrivacyenhancedMD} in the past few years. Preliminary research named Federated Stochastic Gradient Descent (FedSGD) \cite{Chen2016RevisitingDS} is a synchronous approach where a single batch gradient calculation is conducted per round of communication, which is shown to outperform asynchronous approaches. Derived by the idea of FedSGD, Federated Averaging (FedAvg) \cite{McMahan2017CommunicationEfficientLO} adds more computation to each participant by iterating the local update multiple times before the averaging step. It achieves a substantial reduction in the required communication rounds compared with FedSGD when building high-quality models, and thus becomes a typical representative of multi-party learning algorithms.

However, most existing approaches take the form of aggregating parameters \cite{Zhou2019APD} \cite{9082851} \cite{9238427}, such as methods based on averaging local stochastic gradient descent updates for the optimization, to update the global model. Although they could work well empirically under ideal conditions, these approaches defeat the original purpose of the artificial neural networks, i.e., the simulation of human brain cognition, which reduces the interpretability of multi-party learning. Moreover, clients usually generate local data in a not independent or not identically distributed (non-IID) manner, in which aggregating methods come without convergence guarantees and frequently diverge in practical applications \cite{Yang2019FederatedML} \cite{9003490} \cite{9130845}. Besides, an appropriate incentive mechanism that encourages clients to provide high-quality data and communicate more frequently with the server is necessary for the multi-party learning scenario \cite{Khan2020FederatedLF} \cite{Zhan2020ALI}. Additionally, due to the differences of devices in hardware state, battery level and network connectivity in the process \cite{Li2020FederatedLC}, the synchronous scheme is susceptible to unforeseen upload delays in the face of device variability, which results in a decline in the rate of convergence and performance degradation of the global model.

\begin{figure*}[htb!]
  \centering
  \includegraphics[width=0.8\linewidth]{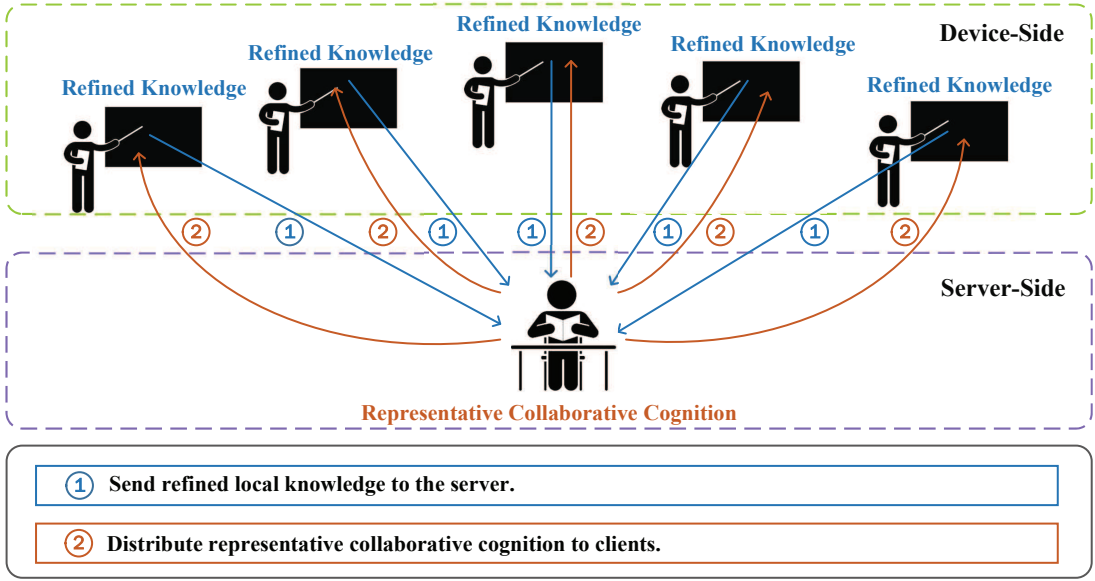}
  \caption{
  Diagram of the many-to-one knowledge sharing framework.}
\label{fig:whole}
\end{figure*}
To address the aforementioned challenges, we present a novel contrastive multi-party learning framework to improve efficiency and effectiveness in an explainable manner, which follows the scenario that humans impart knowledge by extracting key information with distinguishable characteristics instead of sharing the data source directly. Specifically, we employ the modularized design for local models, i.e. cognitive module, descriptive module and the discriminative module, and analogy the multi-party learning scenario as a many-to-one knowledge sharing problem, as depicted in Figure~\ref{fig:whole}. Clients extract the higher-order features of the raw data and refine the representations with contrastive loss, and the distribution is imparted to the server explicitly as the acquired knowledge, which is analyzed synthetically on the server. Afterwards, the server forms a representative collaborative cognition and distributes it to each client for seeking the unity of the individual's cognition in turn. The interactions among clients and the server are more flexible and stimulated by the knowledge aggregation strategy, which follows the principle that contributions correspond with rewards. To the best of our knowledge, this paper is the first to preliminarily explore the potential of explainable knowledge sharing in multi-party learning. The main contributions of this paper are summarized as follows:

\begin{itemize}
	\item{We introduce a contrastive knowledge sharing framework to provide more explainable solutions for multi-party learning problems, and we show that the biomimetic scheme achieves superior performance than conventional parameter aggregation methods in various scenarios.}
	\item{We provide a novel view for handling the problem of statistical heterogeneity (non-IID data). Participants are able to selectively impart the acquired knowledge to the central server, so that the quality of learning can be guaranteed and the global model could avoid being polluted by the untrained parameters.}
  \item{We design an incentive mechanism employed in the presented scheme that enables the global model to demonstrate better performance on more active clients, and it encourages participants to contribute more data with high-quality.}
	\item{The proposed framework achieves superior performance in an asynchronous setting with the assistance of the modularized structure and end-to-end training, and clients could receive immediate feedback for continuous training without requiring bounded-delay assumptions.}
\end{itemize}

The remainder of this paper is organized as follows. Section~\ref{para:2} briefly presents the related backgrounds about multi-party learning and contrastive learning. In Section~\ref{para:3}, we formalize the research problem of the multi-party learning, then the details of our framework are described. Section~\ref{para:4} shows extensive experiments that validate the effectiveness. Finally, we conclude with a discussion of our framework and summarize the future work in Section~\ref{para:5}.

\section{Background and Related Works}

\label{para:2}
\subsection{Multi-Party Learning}
The rapid development of artificial intelligence promotes the demand for a large number of high-quality data. However, data is limited or of poor quality in most fields, making it hard for artificial intelligence algorithms to achieve their optimum performance. Besides the scarcity of high-quality data resources, many areas in reality, such as medicine and finance, also have privacy issues, which become another obstacle to the access to raw data. An effective multi-party learning framework proposed recently \cite{Konecn2016FederatedOD} allows participants to keep their private data and conduct model training locally. Meanwhile, it is models instead of raw data that are aggregated in the server, hence the risk of privacy leakage is minimized.
\begin{figure}[b!]
  \centering
  \includegraphics[width=0.7\linewidth]{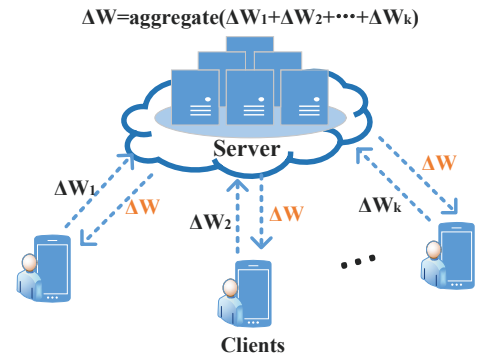}
  \caption{
  Benchmark algorithm: Federated Averaging method.}
\label{fig:fedavg}
\end{figure}

A typical representative of this work is Federated Averaging (FedAvg), which runs several steps of stochastic gradient descent (SGD) in parallel on each client, and then averages the resulting model updates through a central server. Compared with FedSGD and its variants, FedAvg has more local computation and less communication. The gradient update strategy of FedAvg is to calculate the gradient $g_k=\nabla F_k(w_t)$ on $K$ clients, then aggregate these gradients on the central server $\begin{matrix}\sum_{k=1}^K\frac{n_k}{n}g_k \end{matrix}$ and return it to update local models
\begin{equation}
w_{t+1} \gets w_t-\eta\begin{matrix} \sum_{k=1}^K\frac{n_k}{n}g_k \end{matrix},
\end{equation}
as shown in the Figure~\ref{fig:fedavg}.

While alleviating the data privacy problem, multi-party learning algorithms still have several problems to be solved, such as system heterogeneity, statistical heterogeneity, and interpretability. For example, unlike the traditional distributed learning system, multi-party learning cannot control the connection of participants \cite{Lu2020DifferentiallyPA}. When local devices are turned off or the communication network is not available, they have no access to the server and become stragglers, in which conditions the system heterogeneity occurs, and it significantly slows down the convergence under different sampling and averaging schemes \cite{smith2017federated}.

Moreover, the local data on clients is usually non-IID due to different regions or habits of participants \cite{Niknam2020FederatedLF}, which not only makes the theoretical analysis difficult but also challenges the algorithm design. Wang \emph{et~al.} \cite{Wang2019AdaptiveFL} analyzes the convergence bounds of the distributed gradient descent algorithm theoretically, and Stich \cite{Stich2019LocalSC} calculated the local convergence rate of SGD on convex problems. However, they all make some assumptions about the data distribution or equipment condition, i.e. the data are distributed in an IID manner and all devices are well connected, which obviously violates the characteristics of multi-party learning. Lin \emph{et al.} \cite{Lin2020EnsembleDF} investigated ensemble distillation for model fusion, which trains the central classifier through unlabeled data on the outputs of the models from the clients. Sahu \emph{et~al.} \cite{Sahu2020FederatedOI} proposed FedProx to tackle heterogeneity in multi-party networks, which generalizes and re-parameterize the FedAvg. In a word, the statistical heterogeneity is a pressing problem for multi-party learning and could be an insurmountable obstacle for the traditional parameter aggregation methods.

\subsection{Contrastive Learning}
In recent years, the pursuit of higher-level semantic and logical extraction has become the development trend of deep learning, and it is also the essence of contrast learning, which has shown its innovation and potential by learning the similarity and dissimilarity of samples to construct representations. More concretely, contrastive learning can maximize the consistency between different extended views of similar data samples through the contrastive loss in potential space, and it only needs to learn distinguishing features to classify different classes of samples without paying too much attention to fine-grained feature details.

Contrastive learning has been applied to many scenes and has provided excellent performance \cite{Tian2020WhatMF} \cite{Chen2020ASF}. Dai \emph{et al.} \cite{Dai2017ContrastiveLF} introduced contrastive learning to image caption tasks to enhance the uniqueness of the caption description. The Contrastively-trained Structured World Models (C-SWMs) presented in \cite{Kipf2020ContrastiveLO} performs contrastive learning in the environment with a composite structure, which makes the interaction among objects achieve good inductive preference. C-SWMs not only learn interpretable object-based representation, but also overcome the limitations of the pixel reconstruction model. In terms of theoretical analysis of contrastive learning, Arora \emph{et al.} \cite{Arora2019ATA} proposed a theoretical framework for analyzing contrastive learning algorithms by introducing potential classes and hypothetical semantic similarity points from the same potential class. They also proved that the generated representations could reduce the sample complexity of downstream tasks, that is, fewer samples are required for the model to converge.

When handling labeled data, Khosla \emph{et al.} \cite{Khosla2020SupervisedCL} proposed supervised contrastive learning which combines category information with self-supervised contrastive learning. Specifically, supervised contrastive learning leverages category labels to construct positive samples and negative samples, and measures the distance among them. Its training process is to obtain $2N$ samples after 2 kinds of data enhancement for $N$ labeled samples. Then, the backbone network is trained by using labels and loss function to make the representations of the same class close to each other, and make the representations from different classes far away. The loss function is defined as follows:
\begin{equation}
\mathcal{L}_{con}=\sum_{i=1}^{2N}\mathcal{L}_{con}^{(i)},
\label{equ:loss3}
\end{equation}
\begin{equation}
\mathcal{L}_{con}^{(i)}=\frac{-1}{2N_{\tilde{y}_i}-1}\sum_{j=1}^{2N}\textbf{1}_{i\ne j}\cdot \textbf{1}_{\tilde{y}_i=\tilde{y}_j}\cdot \log{\frac{\exp(z_i\cdot z_j/\tau)}{\sum_{k=1}^{2N}\textbf{1}_{i\ne k}\cdot \exp(z_i\cdot z_k/\tau)}},
\label{equ:loss4}
\end{equation}
where $\tilde{y}_i,\tilde{y}_j$ denote the category of sample $i$ and sample $j$, $N_{\tilde{y}_i}$ represents the total number of samples that have the same label as $i$ in a minibatch, and $\tau$ is a scalar parameter greater than 0.

It encourages the encoder to distribute closely aligned representations to all entries from the same class in each instance of Eq.~\ref{equ:loss4}, resulting in a more robust clustering in the latent space \cite{Khosla2020SupervisedCL}. Compared with the cross-entropy loss, the contrastive loss pulls representations of samples from the same class closer in a more natural way instead of forcing them towards a specific target, and the improvement can more vividly simulate the human cognitive and learning scenarios.

\section{Proposed Algorithm}
\label{para:3}
\begin{figure*}[htb!]
  \centering
  \includegraphics[width=0.7\linewidth]{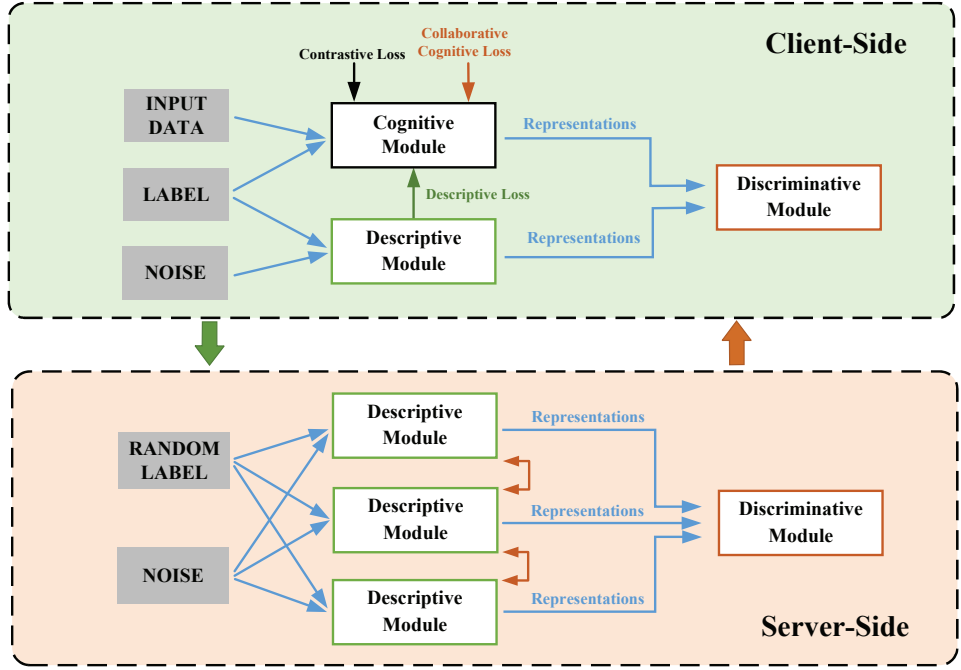}
  \caption{
  A graphical illustration of the contrastive multi-party learning framework, in which blue arrows denote data flow links. The descriptive module in the green box, which masters the higher-order representations of each class of data, is transmitted from clients to the server to teach the central discriminative module in the red box. Then, the latter, along with the collaborative cognitive loss, are sent back to clients for feedback and updates.}
\label{fig:federated}
\end{figure*}

The proposed contrastive multi-party learning framework consists of several clients and a server, in which the former owns private data and learns to describe the distinguishing features of raw data, while the latter could constantly train the discriminant module with knowledge shared by clients, as depicted in Figure~\ref{fig:federated}. In this section, we first give a formal definition of the problem and put forward the properties that the scheme should satisfy. Then we elaborate on the specific work and principles of the three functional modules. Finally, the learning process and relevant supplementation of the framework are presented in detail.

\subsection{Problem Statement}
\label{para:3.1}
Consider there are $K$ clients, each with a local dataset that can only be stored and processed locally, and the goal of multi-party learning is to learn the global statistical model from the decentralized data across a large number of clients. Take non-convex neural network optimization as an example, the scheme should be applicable to the finite-sum objective of the form

\begin{equation}
\underset{{w\in \mathbb{R}^d}}{\rm min}\ f(w)\quad {\rm where}\quad f(w)\overset{\rm def}=\frac{1}{N}\sum\limits_{i=1}^N f_i(w).
\end{equation}

It typically takes $f_i(w)=\ell(x_i,y_i;w)$ for deep learning problems, denoting the loss of the prediction on $N$ samples $(x_i,y_i)$ made with model parameters $w$. Therefore, the objective of multi-party learning with $K$ clients can be rewritten as
\begin{equation}
f(w)=\sum_{k=1}^K \frac{N_k}{N} F_k(w)\ \ {\rm where}\ \ F_k(w)=\frac{1}{N_k}\sum_{i=1}^{N_k}f_i(w),
\end{equation}
which indicates the loss for the global model is formed by the weighted sum of that for all clients, and the weight is associated with the size of the local dataset utilized in the training process. The incentive mechanism enables the global model to demonstrate superior performance on more active clients, and further encourages participants to contribute more more data with high-quality. In particular, given the aforementioned challenges, the multi-party learning framework could be more superior and competitive if it could handle statistical and system heterogeneity in an explainable manner.

\subsection{Functional Module}
\label{para:3.2}
\begin{figure*}[htb!]
  \centering
  \includegraphics[width=0.7\linewidth]{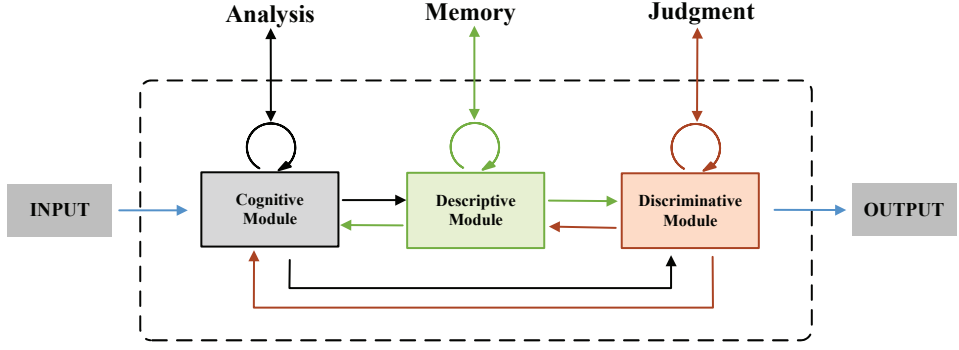}
  \caption{
  The schematic representation of human learning mechanism based on analysis, memory and judgment.}
\label{fig:concept}
\end{figure*}
The proposed scheme works against the traditional parameter aggregation methods since it attempts to simulate the process of human cognition and communication, and the multi-party learning is regarded as a many-to-one knowledge transference problem in a bionic perspective. Naturally, the process of human learning is mainly involved in three modules: cognitive module, descriptive module and discriminative module \cite{Wang2017CognitiveFO}, corresponding to the analysis, memory and judgment for the human brain, as depicted in Fig.~\ref{fig:concept}.

\subsubsection{Cognitive Module}
\label{para:3.2.1}
The function of the cognitive module is to extract high-order dense representations from the raw data for the subsequent judgment. Specifically, there is no need to pay attention to every detail of samples, as long as the learned features can distinguish the sample from the others. Taking a dollar bill as an illustrative example, despite having seen it countless times, it would make a big difference if one tries to draw a bill from memory \cite{anand2020contrastive}. Therefore, the cognitive module should summarize more clear and concise knowledge instead of bloated and redundant ones. The cognitive module is always instantiated as an encoder, whose network structure can be designed flexibly according to the type and characteristics of the input data, e.g. convolutional layers \cite{Sulam2020OnMB} for image data and recurrent layers \cite{Wang2019SalientOD} for text data. In the training stage of the encoder, the contrastive loss is introduced, which is defined as follows:
\begin{equation}
\begin{split}
\mathcal{L}_{con}(z^{(i)},z^{(j)})&=\textbf{1}_{y^{(i)}=y^{(j)}}{\lVert z^{(i)}-z^{(j)}\rVert}_2\\
&+\textbf{1}_{y^{(i)}\ne y^{(j)}}\max(0,M-{\lVert z^{(i)}-z^{(j)}\rVert}_2).
\label{equ:loss5}
\end{split}
\end{equation}

The contrastive loss function is calculated on pairwise output embedding vectors $(z^{(i)},z^{(j)})$. It essentially retrogrades to the Euclidean distance if the two vectors have the same label, i.e. $y^{(i)}=y^{(j)}$, and is otherwise equivalent to the hinge loss. The margin parameter $M>0$ is leveraged to impose a lower bound on the distance between a pair of samples with different labels. Intuitively, it encourages representations with the same labels closer to each other, while those with different labels are enforced to form a margin $m$ among them, thus the encoder is able to capture distinctive features of each class for downstream memory and judgment.

It is worth noting that representations of each class for a single client is assumed to be distributed in an IID manner, though there could exist statistical heterogeneity among different clients. Further, due to the contrastive loss that imposes restrictions on the distribution of embedding vectors, their mathematical expectations and variances are finite, and it has been proved that these dense representations follow the multivariate Gaussian distribution $\mathcal{N}(\mu, \Sigma)$ according to the Lindeberg{-}L\'evy Central Limit Theorem \cite{Mena2019StatisticalBF}. Therefore, the representative collaborative cognition $\mathcal{R}_j$ for each class $j$ acquired by a client can be summarized by the mathematical expectation $\mu$ and the covariance matrix $\Sigma$, i.e.,

\begin{gather}
\mu = \frac{1}{m} \sum_{i=1}^m z^{(i)}, \\
\Sigma = \frac{1}{m} \sum_{i=1}^m (z^{(i)}-\mu)(z^{(i)}-\mu)^T,
\end{gather}
in which $z^{(i)}$ is an embedding vector, and $m$ means the number of vectors in class $c$. Note that we have $\Sigma\leftarrow\Sigma+\gamma I$ where $\gamma$ is a small number and $I$ is the identify matrix, hence $\Sigma$ is guaranteed to be invertible to facilitate subsequent calculations.

However, as mentioned before, due to the permutation invariance of hidden elements (such as channels for convolution layers) in neural networks, there could exist various cognition for the same class in different clients. If one develops a cognitive system without acquiring common knowledge from the outside world, there will be cognitive bias in descriptions of objects for the person, e.g. deems green as blue and that square is round, hence the knowledge sharing process becomes baffling and counteractive. Therefore, cognitions of all clients need to be unified by introducing a collaborative cognitive loss into cognitive modules, the calculation of which is given in Section~\ref{para:3.3}. 
\begin{equation}
\begin{split}
W_2(\mathcal{N}(\mu_c, \Sigma_c);&\ \mathcal{N}(\mu_\mathcal{R}, \Sigma_\mathcal{R})) = ||\mu_c-\mu_\mathcal{R}||_2^2\\
&+tr(\Sigma_c+\Sigma_\mathcal{R}-2(\Sigma_c^{1/2} \Sigma_\mathcal{R} \Sigma_c^{1/2})^{1/2}),
\end{split}
\end{equation}
which is the second-order Wasserstein distance between the local knowledge and common knowledge. $\mu_c$, $\Sigma_c$ represent client's local cognitions for class $c$, and $\mu_\mathcal{R}$, $\Sigma_\mathcal{R}$ denote representative collaborative cognition. As both covariance matrices are symmetric positive definite matrices, it satisfies $tr(\Sigma_c^{1/2} \Sigma_\mathcal{R} \Sigma_c^{1/2})=tr(\Sigma_c \Sigma_\mathcal{R})$ in the commutative case, and we have
\begin{equation}
\begin{split}
tr(\Sigma_c+\Sigma_\mathcal{R}&-2(\Sigma_c^{1/2} \Sigma_\mathcal{R} \Sigma_c^{1/2})^{1/2})\\
&=tr(\Sigma_c)+tr(\Sigma_\mathcal{R})-2(tr(\Sigma_c^{1/2} \Sigma_\mathcal{R} \Sigma_c^{1/2})^{1/2})\\
&=tr((\Sigma_c^{1/2}-\Sigma_\mathcal{R}^{1/2})^2)\\
&=||\Sigma_c^{1/2}-\Sigma_\mathcal{R}^{1/2}||^2_F,
\end{split}
\end{equation}
in which $||\cdot||_F$ is the Frobenius norm of the matrix. The collaborative cognitive loss $\mathcal{L}_{col}$ for a class $c$ can be thus defined as:
\begin{equation}
\mathcal{L}_{col} = ||\mu_c-\mu_\mathcal{R}||_2^2+||\Sigma_c^{1/2}-\Sigma_\mathcal{R}^{1/2}||^2_F.
\label{loss13}
\end{equation}
It imposes common knowledge constraints that the server approves, as depicted by the brown arrows in Figure~\ref{fig:federated}, in order to force generated embedding vectors of each client's cognitive module to follow similar distribution to the others. The distribution can be continuously corrected by local data, and the modification is transmitted to the server by the descriptive module. The final loss of the cognitive module consists of the above two parts:

\begin{equation}
\mathcal{L}_{cog} = \mathcal{L}_{con}+\mathcal{L}_{col}.
\label{equ:loss7}
\end{equation}

\subsubsection{Descriptive Module}
\label{para:3.2.2}
The descriptive module aims to imitate features reported by the cognitive module and impart the acquired knowledge to the central discriminative module on the server. After the cognitive module extracts representative features from the raw data, the descriptive module tries to generate vectors that follow the same distribution as these features from input Gaussian noises and corresponding labels. The Mahalanobis Distance \cite{Ye2017LearningMD}, which considers the relationship among different features, is leveraged to measure the difference between vectors generated by the two modules, and the descriptive module is updated by descending
\begin{equation}
\mathcal{L}_{des}=\sum_{i=1}^{N}\sqrt{(z^{(i)}_{cog}-z^{(i)}_{des})^T\Sigma^{-1}(z^{(i)}_{cog}-z^{(i)}_{des})},
\label{equ:loss8}
\end{equation}
in which $N$ is the batch size, and $\Sigma$ is the local covariance matrix associated with the class of $z^{(i)}$. The descriptive module maps the standard Gaussian distribution to that of representative features, and it is capable of reporting representative features to the server to update the central discriminative module. It prevents privacy disclosure of the raw data against various inversion attacks \cite{Zhu2019DeepLF}, which can merely derive noise data from the parameters or gradients of the descriptive module. Therefore, the modular design is safer compared with conventional parameter aggregation algorithms when facing a curious or semi-honest server.

\subsubsection{Discriminative Module}
\label{para:3.2.3}
The function of the discriminative module is to predict labels of representations from the cognitive module and descriptive module, which is instantiated as a classifier. The classification accuracy of the discriminative module intuitively reflects the quality of features extracted by the cognitive module and knowledge learned by the descriptive module.

The cross-entropy loss is utilized for the optimization of the discriminative module, and it is able to update the three networks simultaneously by descending two losses defined as follows:
\begin{equation}
\mathcal{L}_{cc}=-\sum_{i=1}^{N}\sum_{c=1}^{C}y_c^{(i)}\log{y_{cog_c}^{(i)}},
\label{equ:loss9}
\end{equation}

\begin{equation}
\mathcal{L}_{cd}=-\sum_{i=1}^{N}\sum_{c=1}^{C}y_c^{(i)}\log{y_{des_c}^{(i)}},
\label{equ:loss10}
\end{equation}

where $N$ is the batch size, $y_c$ is the real labels of samples, $y_{cog_c}$ indicates the probability that embedding vectors generated by the cognitive module belong to the $c$-th class, and $y_{des_c}$ is that for the descriptive module, and $C$ means the number of classes, i.e. the output dimension of the discriminative module. The total loss of the discriminative module $\mathcal{L}_{dis}$ is obtained by weighting the predicted results of samples from the cognitive module and descriptive module with a scalar $\alpha$ in the following equation:

\begin{equation}
\mathcal{L}_{dis}=\alpha\mathcal{L}_{cc}+(1-\alpha)\mathcal{L}_{cd}.
\label{equ:loss11}
\end{equation}

\subsection{Training Process}
\label{para:3.3}
The training process of the proposed framework consists of a client-side local training stage and a server-side knowledge sharing stage. Each client is able to flexibly schedule the execution of the two alternate phases. In other words, they can perform training and upload at their own convenience without the restriction of synchronizing with other clients.

Specifically, the update process on local clients is depicted in Algorithm~\ref{alg:client}. Once receiving the central discriminative module (classifier) $\mathcal{M}_C$ that has the ability of global judgment from the server, each client replaces its local discriminative module with $\mathcal{M}_C$ and splits local data into batches $\mathcal{B}$ of minibatch size $B$. It calculates the contrastive loss and collaborative cognitive loss at first, then updates the local encoder $\mathcal{M}_E$. When the encoder has been able to capture distinctive features of each class and achieves cognitive unification with other clients, the descriptive loss is computed and the local generator $\mathcal{M}_G$ can be updated, thus more comprehensive knowledge is reserved in the network parameters of $\mathcal{M}_G$. Afterwards, the extracted features serve as the input of the classifier $\mathcal{M}_C$, and fine-tuning is conducted on all modules to reduce classification error. After a training epoch, the descriptive module, which possesses the knowledge of high-order features of local data, is sent to the server to impart the feature distribution to the central discriminative module.

\begin{algorithm}[!t]
  \caption{ Clients Update Procedure}
  \begin{algorithmic}[1]
  \label{alg:client}
  \REQUIRE
  Central discriminative module $\mathcal{M}_C$; collaborative cognition $\mu$, $\Sigma$
  \ENSURE
  Descriptive module $\mathcal{M}_G$

  \textbf{Client executes:}\\
  \STATE Replace the local discriminative module with $\mathcal{M}_C$;\\
  \STATE Split local data into batches $\mathcal{B}$ of minibatch size $B$;\\

  \FOR{batch $b \in \mathcal{B}$}
     \STATE Calculate the contrastive loss $\mathcal{L}_{con}$ and collaborative cognitive loss $\mathcal{L}_{col}$ according to Eq.~\ref{equ:loss5} and Eq.~\ref{loss13};
      \STATE Update encoder $\mathcal{M}_E$;
      \STATE Calculate the descriptive loss $\mathcal{L}_{des}$ according to Eq.~\ref{equ:loss8};
      \STATE Update generator $\mathcal{M}_G$;
      \STATE Calculate the discriminative loss $\mathcal{L}_{dis}$ using Eq.~\ref{equ:loss11};
      \STATE Update classifier $\mathcal{M}_C$ and the upstream modules $\mathcal{M}_E$, $\mathcal{M}_G$;
      
  \ENDFOR
  \RETURN descriptive module $\mathcal{M}_G$ to the server
  \end{algorithmic}
\end{algorithm}
Moreover, the synchronous knowledge sharing process on the server-side is shown in Algorithm~\ref{alg:server}, but note that the scheme is capable of working in an asynchronous manner without any modification. After receiving descriptive module $M_G^k$ from client $k$, the server feeds $M_G^k$ with noise and labels to generate data embeddings that follow the same distribution as client $k$, hence the central discriminative module $\mathcal{M}_C$ performs incremental learning on the embeddings. Considering that raw data in some clients could be insufficient or of poor quality, it is reflected by a large variance, that is, the client randomly generates mapping vectors for this class of data without refined extraction. Therefore, in order to prevent $\mathcal{M}_C$ from being disturbed by such data, the trace of the covariance matrix $\Sigma_k$ for mapping vectors, which can be interpreted as the dispersion of vectors, i.e. confidence in the acquired knowledge, is calculated and compared with that of the representative collaborative cognition $\Sigma_\mathcal{R}$. The incremental learning is conducted only in the condition that
\begin{equation}
tr(\Sigma_k)<\beta\cdot tr(\Sigma_\mathcal{R}),
\end{equation}
which indicates that client $k$ is believed to describe the data correctly at confidence level $\beta$. Meanwhile, the representative collaborative cognition $\mu$ and $\Sigma$ for class $j$ is updated during the process following the winner-take-all principle:

\begin{equation}
\left\{
  \begin{array}{lr}
  \Sigma_\mathcal{R}\leftarrow\Sigma_k, &  \\
  \\
  \mu_\mathcal{R}\leftarrow\frac{1}{2}(\mu_k+\mu_\mathcal{R}), &
  \end{array}
\right.
{\rm if}\ tr(\Sigma_k)<\frac{1}{\beta}\cdot tr(\Sigma_\mathcal{R})
\label{equ:update}
\end{equation}
where the trace is utilized to measure the quality of the description, and cognition that has minimum within-class scatter is regarded as the representative collaborative one. Otherwise, collaborative cognition remains the same, while the discriminator continues to be updated and assumes the role of knowledge integration. All clients could reach a consensus to reduce cognitive dissonance, so that the central server will not be misled. Afterwards, the central discriminative module $\mathcal{M}_C$ and representative collaborative cognition $\{\mu_c, \Sigma_c\}_{c\in C}$ are sent back to client $k$ to integrate global knowledge for further training. 

The modular deployments and knowledge transfer mechanisms improve the interpretability of multi-party learning by imitating various functional zones of the human brain. When meeting the challenge of statistical heterogeneity, the discriminative module could avoid being polluted by unlearned classes that randomly generate vectors, which usually accompanied by large diagonal elements and can be filtered out in the incremental learning stage. Besides, the many-to-one teaching system naturally supports asynchronous communication since there is no parameter aggregation, and clients have the flexibility to join the learning process at a time convenient to them. Moreover, clients with better cognition and description of data will lead the representative collaborative cognition, and those who participate in interaction actively will facilitate the personalized adaptation of the central discriminative module to them, thus attracting and motivating more clients to participate and contribute their own data.

\begin{algorithm}[!t]
  \caption{ Contrastive Knowledge Sharing Framework}
  \begin{algorithmic}[1]
  \label{alg:server}
  \REQUIRE
  Communication rounds $T$; $K$ clients indexed by $k$
  \ENSURE
  Central discriminative module $\mathcal{M}_C$

  \textbf{Server executes:}

  \STATE Initialize the same local modules for all clients;\\

  \FOR{each round $t$=1,2,...T}
     \STATE Send central discriminative module $\mathcal{M}_C$ and collaborative cognition $\mu$, $\Sigma$ to each client;
     \FOR {each client $k$ \textbf{in parallel}}
      \STATE Receive $\mathcal{M}_G^k$ from local client's update;
      \ENDFOR
     \STATE Generate embeddings by $\mathcal{M}_G$;
     \STATE Update representative collaborative cognition $\mu$, $\Sigma$ according to Eq.~\ref{equ:update};
     \STATE Update $\mathcal{M}_C$ with generated embeddings;
      
  \ENDFOR
  \RETURN Central discriminative module $\mathcal{M}_C$
  \end{algorithmic}
\end{algorithm}

\section{Experiments}
\label{para:4}

\subsection{Dataset}
The MNIST dataset \cite{deng2012mnist} consists of 0-9 handwritten digits, with 60000 handwritten digits in the training set and 10000 in the test set. The test set came from the original NIST program. Each data unit is a $28\times28$ grayscale image with single channel.

The CIFAR-10 dataset \cite{alex:2009} is a small color image data set closer to real objects, with 50000 training images and 10000 test images. It contains 10 classes of RGB color images, each image size is $32\times32$, each category has 6000 images.

The 20 Newsgroups dataset \cite{Lang95} is one of the international standard datasets for text classification, text mining and information retrieval. The dataset collects about 20000 newsgroup documents, which are evenly divided into 20 newsgroup collections with different topics. Some newsgroups have similar themes, while others are completely irrelevant.


The Stanford Sentiment Treebank (SST-5) dataset \cite{Lei2018SentimentLE} contains 9646 training samples and 2210 test samples, where each one is annotated as \emph{very negative, negative, neutral, positive}, or \emph{very positive}.

The FEMNIST dataset \cite{Caldas2018LEAFAB} is a 62-class image classification dataset. It is built by partitioning the digit or character in Extended MNIST dataset based on the writer.

\subsection{Experimental Settings}
The experiments investigate the testing accuracy and convergence speed, and these measures are derived from different methods for comparison. The centralized learning with mini-batch gradient descent serves as the baseline, which gathers all data in the server and enables the model to train on it without privacy concerns. Besides, the classic multi-party learning algorithm, i.e., the Federated Averaging (FedAvg) \cite{McMahan2017CommunicationEfficientLO}, is compared with the proposed scheme as well.

We build a learning scenario with 20 clients and all of them participate in each communication round. The number of local samples is set to be 800 (some repeated sentences for SST-5), except that the number of local samples for CIFAR-10 is 1600 as such leading to a more accurate model. The local batch size is 64 for all modules and comparison algorithms. Moreover, the number of samples generated on the server is 800 for each client as well. Other key impacting factors are weight scalar $\alpha$ for the loss of discriminative module and the confidence level $\beta$ for knowledge updating on the server. The former is set 0.9 due to the faster convergence speed of the descriptive module; and it is found that the model achieves optimal performance when $\beta$ = 1.25 is taken through extensive experiments.

Since functions of the discriminative module (classifier) descriptive module (generator) and are relatively simple, they are instantiated as a multi-layer perceptron with three and four fully-connected layers, respectively. As mentioned before, the network structure of the cognitive module (encoder) can be designed flexibly according to the type and characteristics of the input data. Hence, it consists of three $3\times3$ convolutional layers with (64, 128, 256) channels for MNIST and five $3\times3$ convolutional layers with (64, 192, 384, 256, 256) channels for CIFAR-10, and an average pooling layer is utilized for mapping each feature map to a single value, which means the output of the discriminative is a vector of 256 dimensions. For text datasets, the encoder has an embedding layer of 32 dimensions and the max length is 25, which follows a convolutional layer with 64 channels and the kernel size is $4\times32$. A bi-directional recurrent layer with 128 hidden unit is stacked, and the output is flattened in the top layer.

For the baseline and FedAvg, their network structure is the chained connection of the encoder and classifier, i.e. a common classification network, which follows the same design of ours for fairness. In order to evaluate the effectiveness of the proposed scheme when handling the challenge of statistical heterogeneity, system heterogeneity and closed-loop characteristics, extensive experiments are conducted on IID data and non-IID data, and the framework in the asynchronous setting is compared with that in the synchronous scenario. In addition, how the incentive mechanism works is demonstrated as well. Each experiment is repeated for 5 trials and the averages are reported.

\subsection{Results on IID Data}
\begin{figure*}[!t]
  \centering
  \subfigure[MNIST]{
    \includegraphics[width=0.234\linewidth]{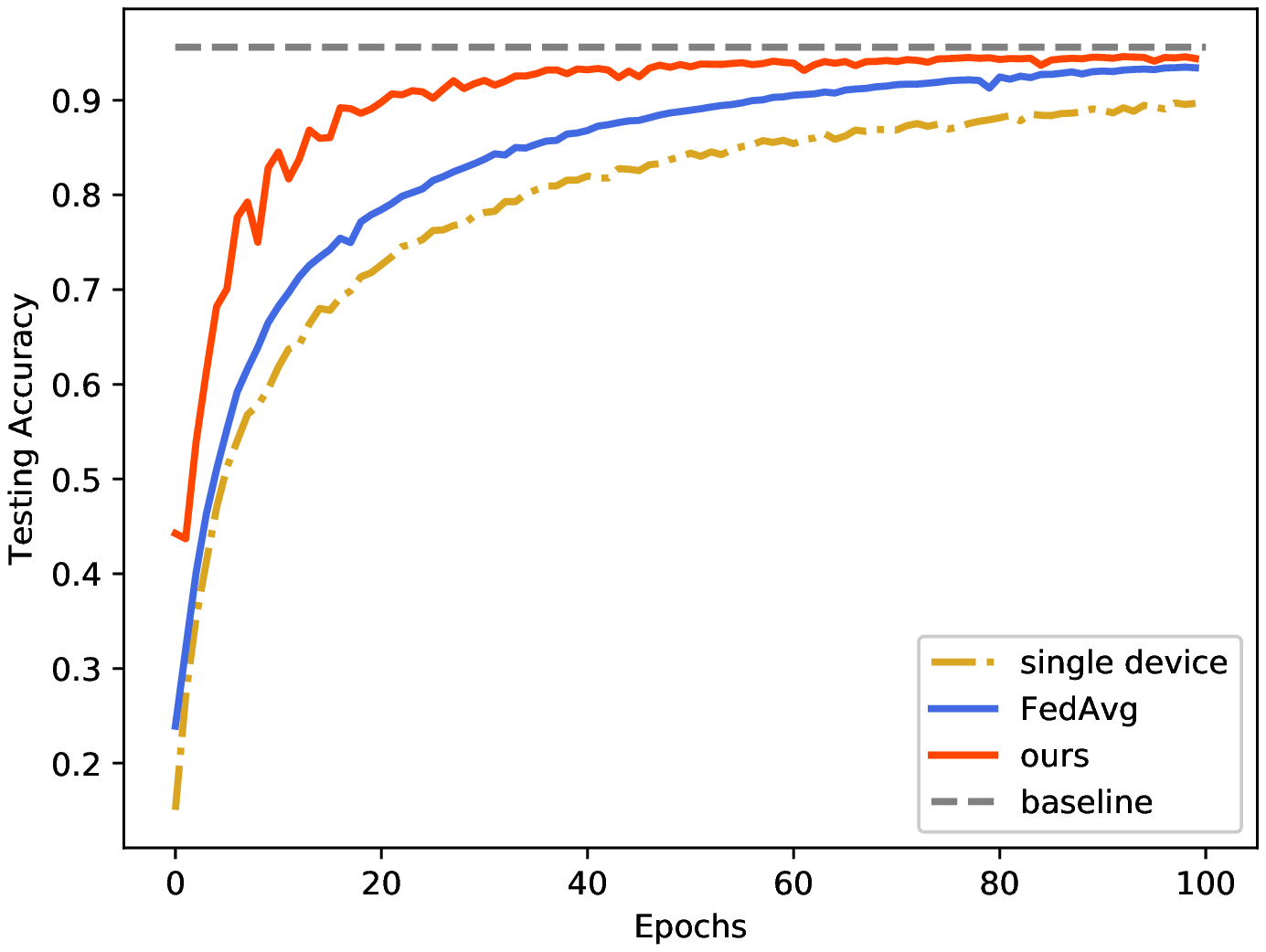}}
  \subfigure[CIFAR-10]{
    \includegraphics[width=0.234\linewidth]{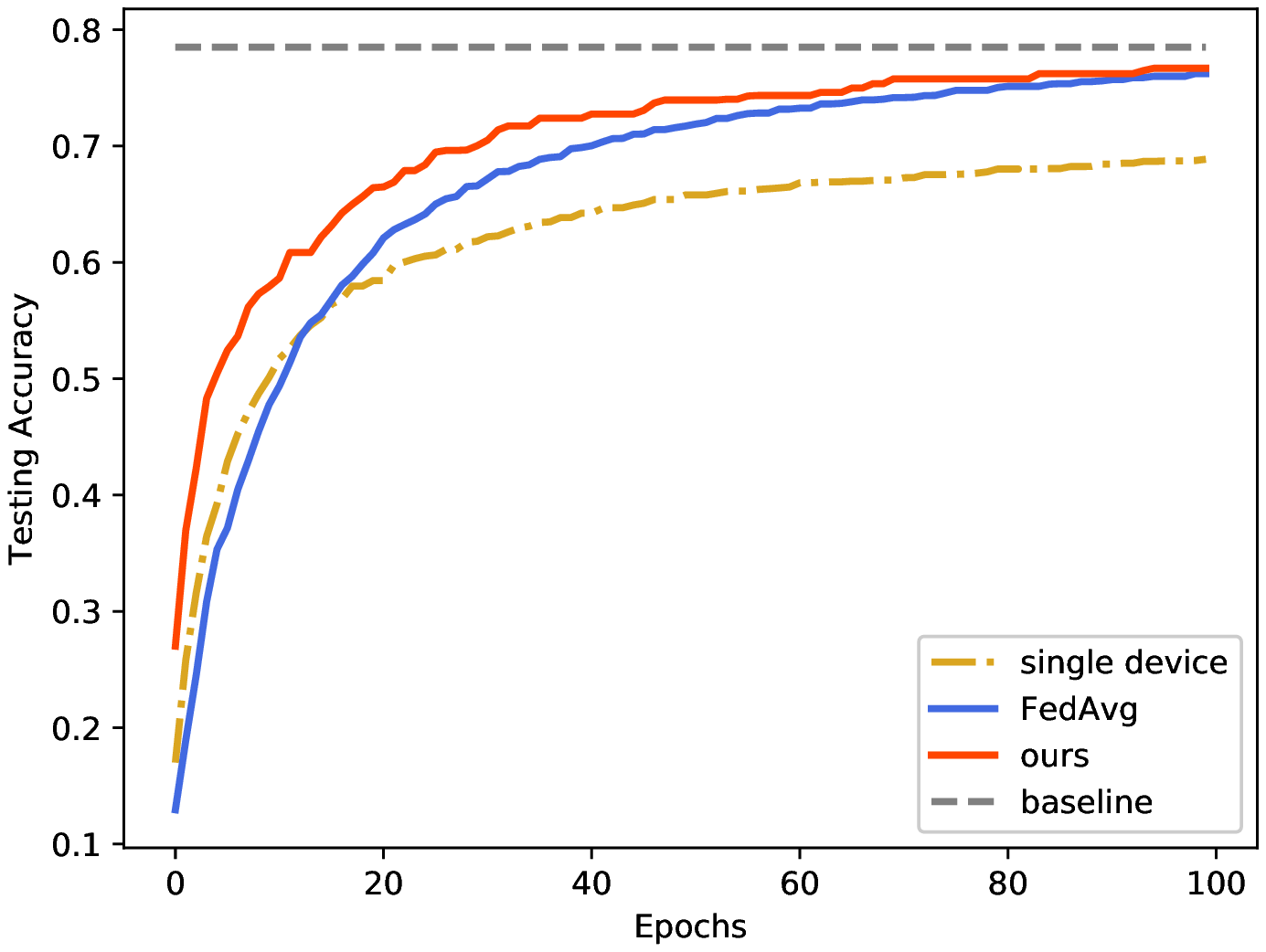}}
  \subfigure[20 Newsgroup]{
    \includegraphics[width=0.238\linewidth]{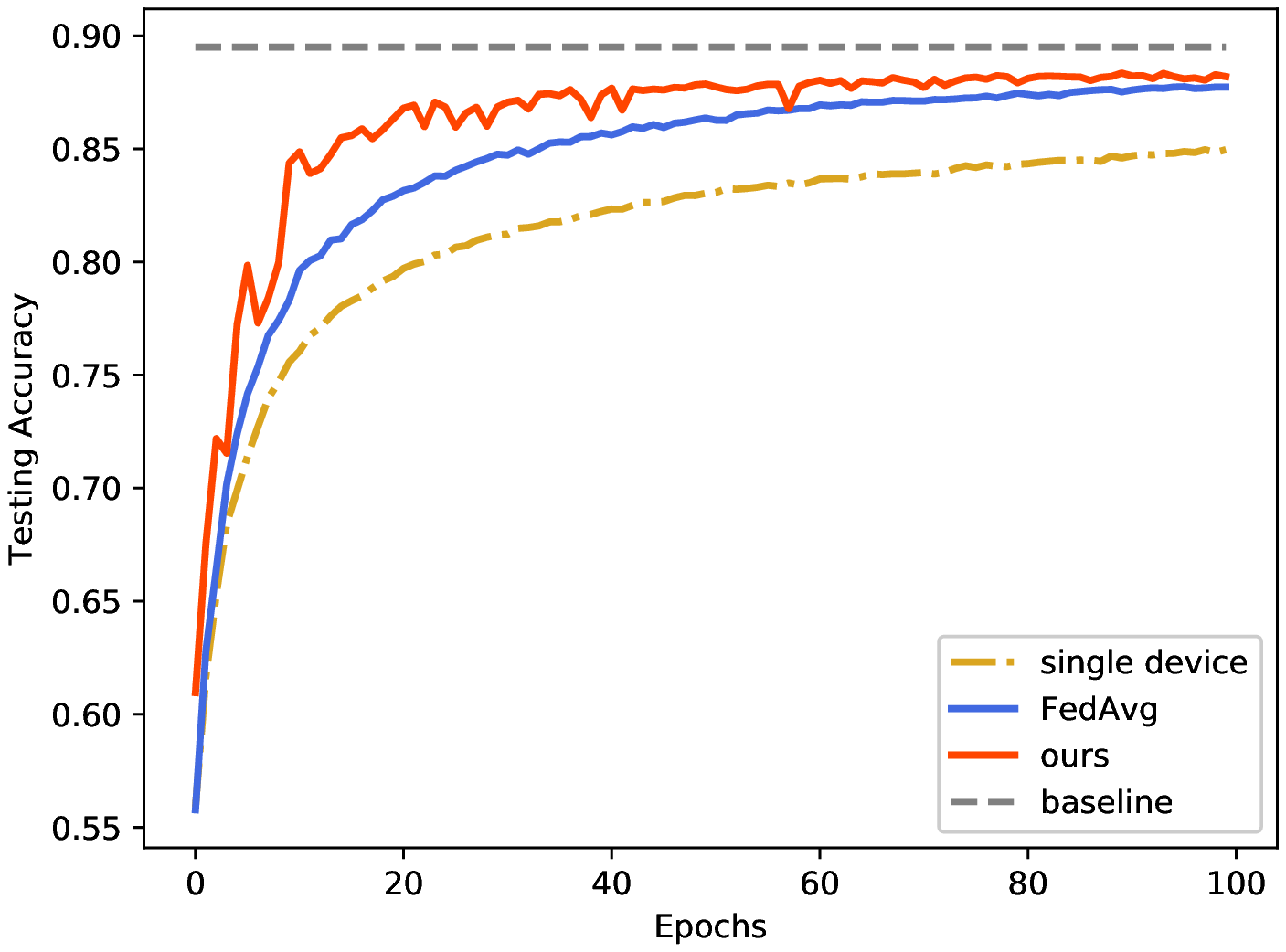}}
  \subfigure[SST-5]{
    \includegraphics[width=0.238\linewidth]{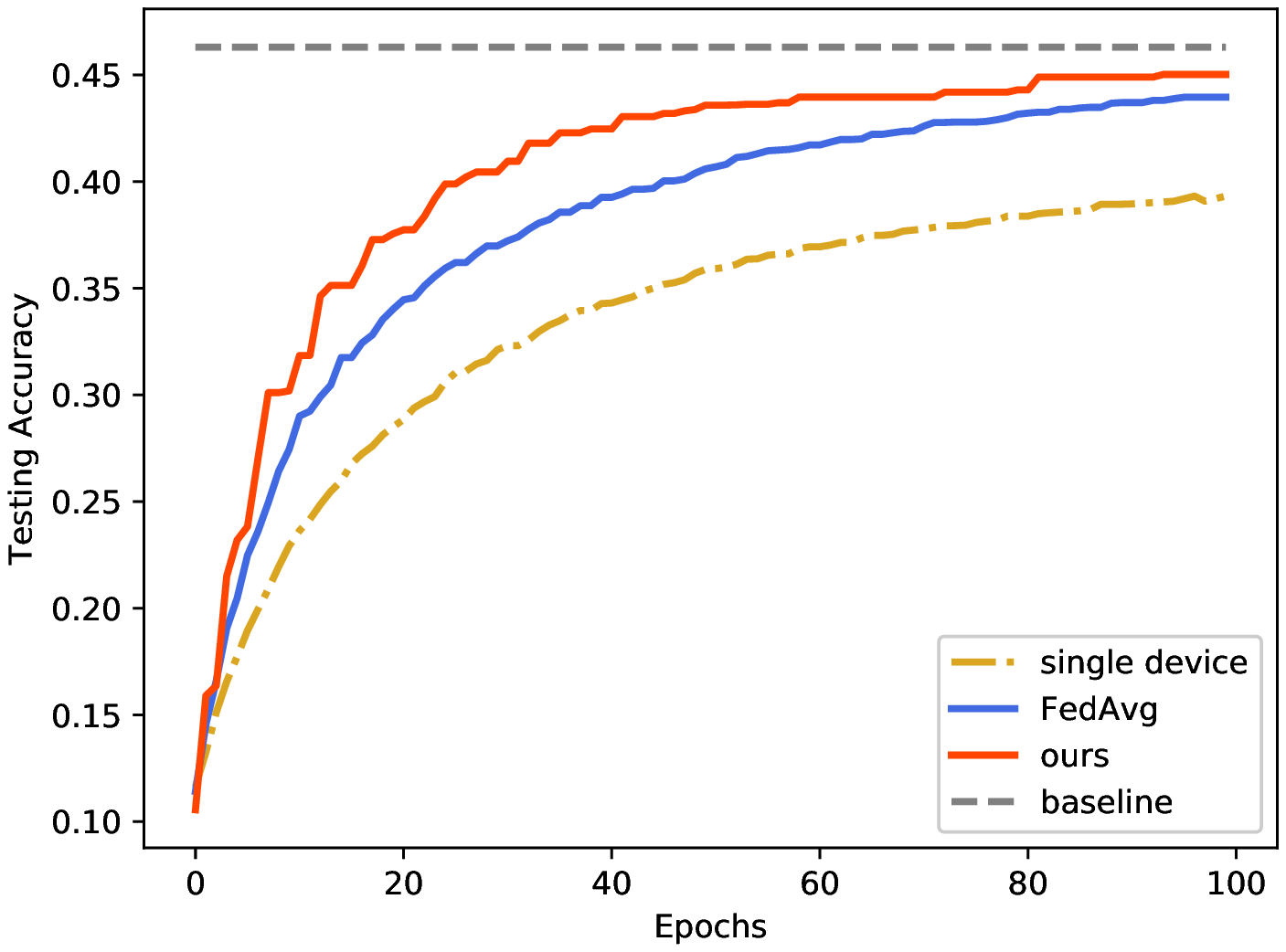}}
  \caption{Testing accuracy on four datasets in the IID setting.}
\label{fig:iid}
\end{figure*}

Since the explainable knowledge sharing framework provides new solutions for multi-party learning problems, its performance is evaluated in various scenarios and compared with the benchmarking, i.e. the conventional FedAvg algorithm, and the testing accuracy on four datasets in the IID setting are given in Fig.~\ref{fig:iid}. To investigate the influence of different aggregation rules of FedAvg and ours, we introduce the comparative studies on a single device, which follows the same hyper-parameter and data distribution, except that it trains on its own data and does not participate in multi-party learning. Note that in the single device condition, where devices are trained merely on their local dataset without aggregation, only one in two algorithms is drawn for simplicity, as the model structure and experimental settings are consistent, and they show similar performance without aggregation.
\begin{figure*}[!t]
  \centering
  \subfigure[MNIST]{
    \includegraphics[width=0.235\linewidth]{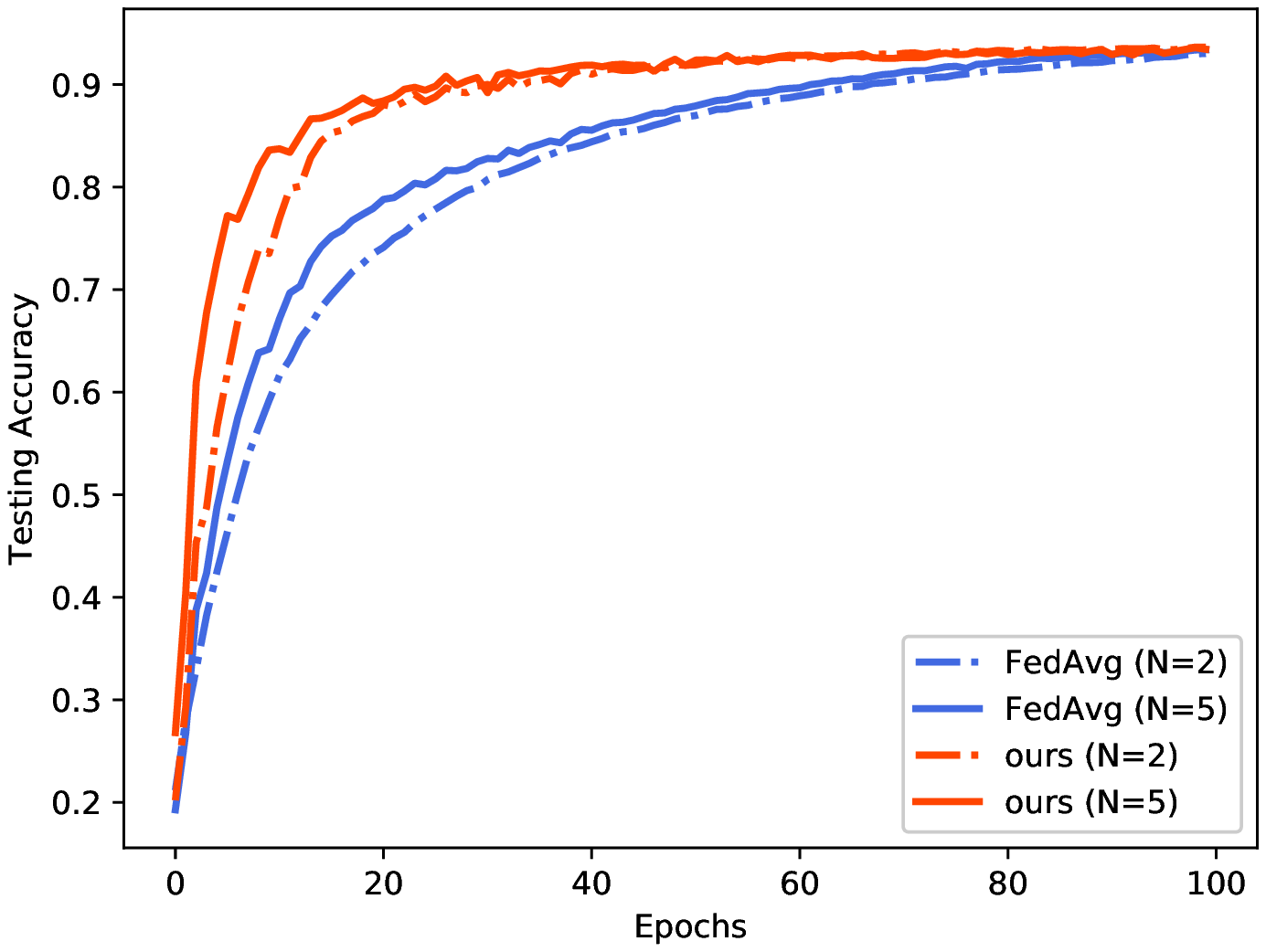}}
  \subfigure[CIFAR-10]{
    \includegraphics[width=0.235\linewidth]{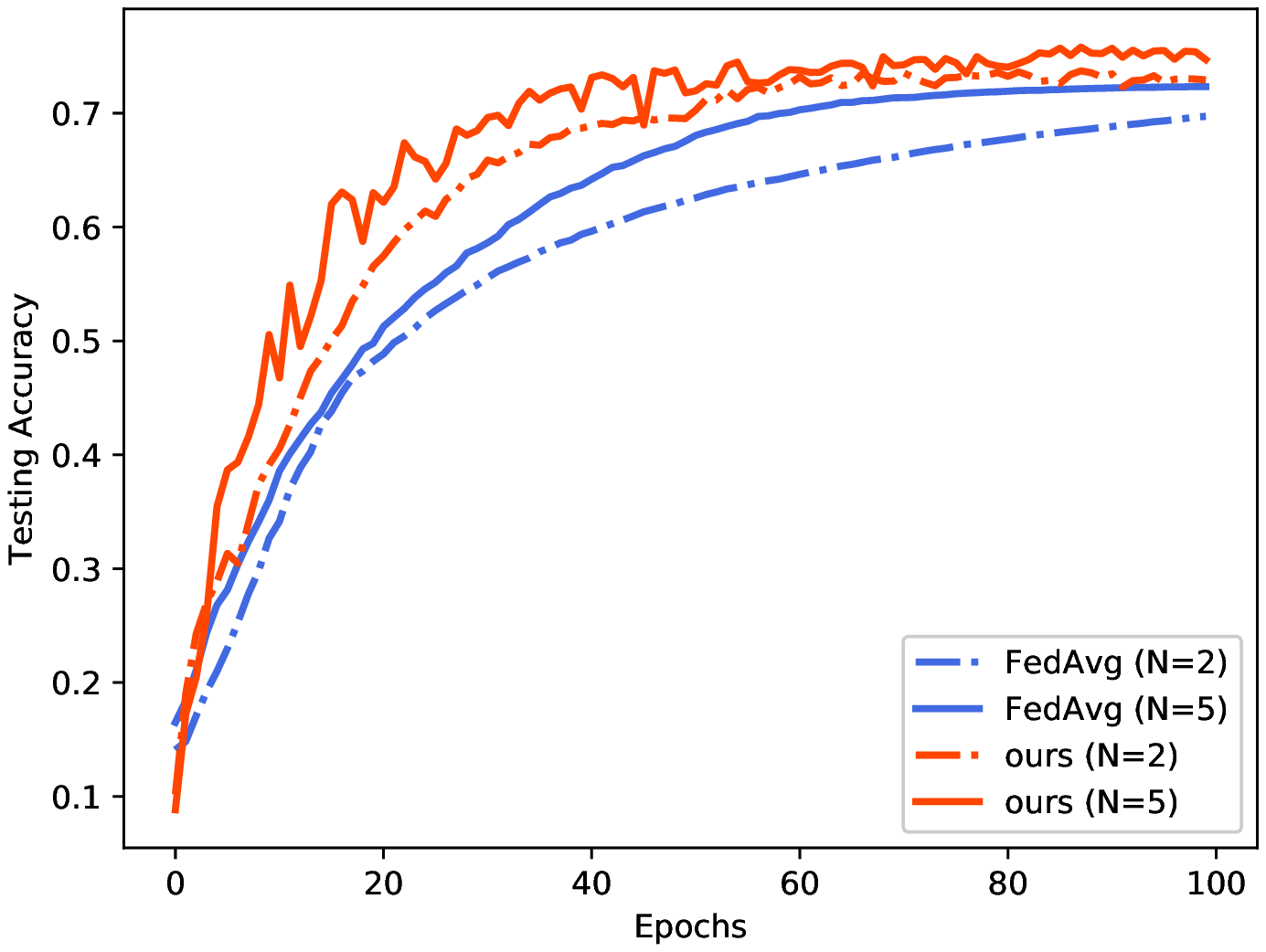}}
  \subfigure[20 Newsgroup]{
    \includegraphics[width=0.235\linewidth]{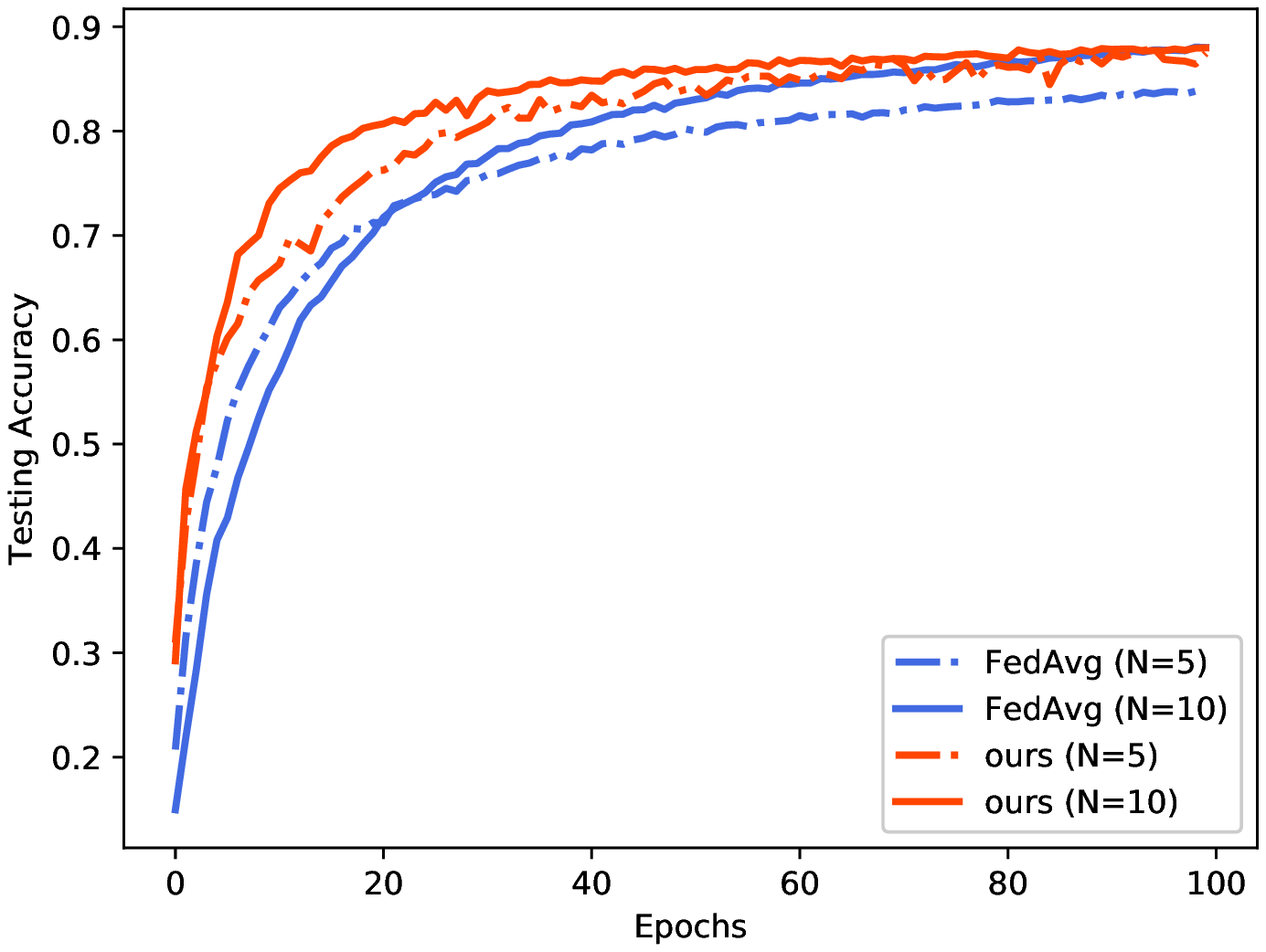}}
  \subfigure[SST-5]{
    \includegraphics[width=0.24\linewidth]{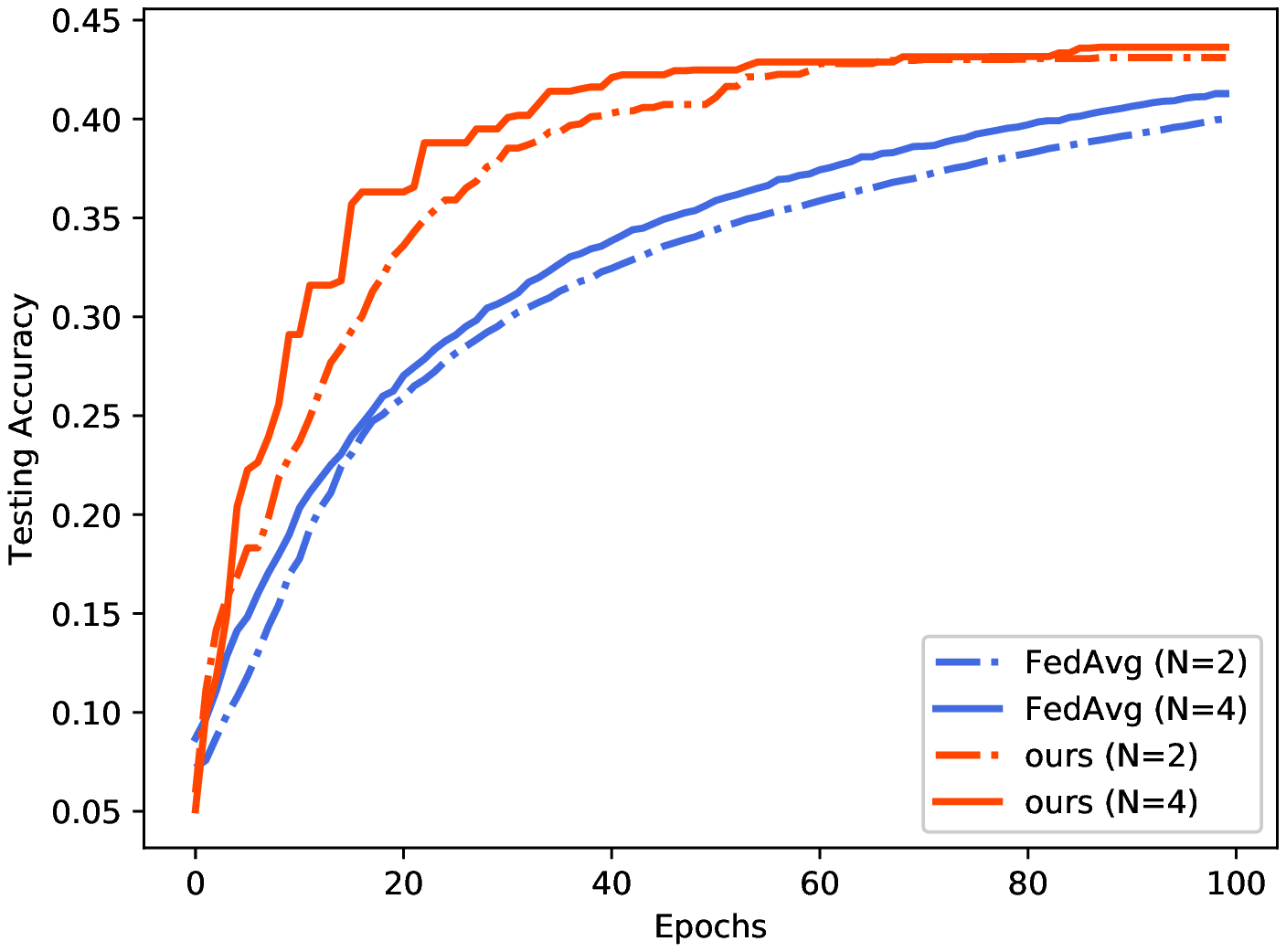}}
  \caption{Testing accuracy on four datasets in the non-IID setting.}
\label{fig:noniid}
\end{figure*}

Compared with the single device, both aggregation rules, i.e. the parameter averaging and knowledge sharing, demonstrate considerable performance improvement, especially on datasets with more complicated data distribution, and our approach even achieves competitive performance against the jointly trained baseline on the MNIST dataset. Though FedAvg achieves a testing accuracy close to that of ours, the proposed scheme converges faster with the assistance of the knowledge sharing strategy. Due to the permutation invariance of internal elements (e.g. channels for convolution layers and hidden elements for recurrent layers) in neural networks, these variants of layers that only differ in the ordering of parameters will ruin the learned feature extraction signatures, which reduces the convergence rate of the global model in parameter averaging approaches. On the contrary, our method is performed following the learning rules of neural networks without parameter aggregation, and the cognitive differences are constrained by the representative collaborative loss; thus the model efficiency and framework interpretability are improved.

\subsection{Results on Non-IID Data}

For the challenge of statistical heterogeneity in multi-party learning, we simulate the scene that there only exist a few categories in the local training stage for clients, in order to demonstrate the effectiveness of our approach. To be specific, we randomly assign several classes of data to each client and keep the number of local samples unchanged. Since the data is partitioned and trained in the non-IID setting, the baseline that gathers all data in the server is not introduced. In addition, the influence of the number of categories $N$ deployed in clients is also measured in the experiments. For MNIST and CIFAR-10 that own 10 classes of data, $N$ is set 2 and 5 in the training stage. As there exist more categories in the 20 Newsgroups dataset, the value of $N$ is 5 and 10, while it equals 2 and 4 for the SST-5 dataset with 5 categories. When calculating the testing accuracy, the test set has all classes of data and follow the IID distribution, and the same setting is employed on FevAvg.

It is observed that the global model converges faster with a larger number of categories on each client for both the FedAvg method and ours. Besides, similar to the performance in the IID scenario, the proposed method obtains superior performance against FevAvg in the model convergence speed and the accuracy. Participants are able to selectively impart the acquired knowledge to the central server explicitly, whose quality is further evaluated according to the dispersion on the server. Therefore, the quality of learning can be guaranteed and the global model could avoid being polluted by redundant information or untrained parameters. Our knowledge sharing framework usually achieves the optimal performance within 40 rounds of communication on four datasets, while the FedAvg algorithm is still not convergent at the end of the training, especially on the SST-5 dataset that is hard to train.

Experiments on data with both IID and non-IID distribution demonstrate the superiority of our knowledge sharing scheme than parameter averaging approach. The proposed method achieves significant accuracy and convergence speed gains on image and text datasets with various networks, though it has a larger training variance compared with FedAvg. The curves fluctuate a lot since updates of the collaborative cognition will affect the performance of a subset of clients by slightly changing their local optimal, while the collaborative cognition tends to be unified along with the iteration. Meanwhile, the privacy is well preserved because the modules directly associated with private data are always kept locally, and the server cannot derive information of the raw data from parameters or gradients of the descriptive module, whereas it is a potential risk for the FedAvg algorithm and its variants \cite{shokri2015privacy} \cite{abadi2016deep}.

\begin{table*}[!t]
\small
\renewcommand{\arraystretch}{1.3}
  \centering
  \caption{Testing accuracy of different multi-party learning algorithms.}
  \setlength{\tabcolsep}{4.7mm}{
    \begin{tabular}{cccccc}
    \hline
      \multirow{2}{*}{\bf{Methods}}&\multicolumn{5}{c}{\bf{Stragglers (\%)}}\\
      \cline{2-6}
      &0&20&40&60&80\\
      \hline
      FedAvg&81.81$\pm$1.39&79.33$\pm$1.92&76.89$\pm$2.63&74.05$\pm$2.74&71.10$\pm$2.97\\
      FedProx&82.96$\pm$1.22&82.20$\pm$1.49&81.32$\pm$1.62&80.34$\pm$1.97&79.70$\pm$2.14\\
      FedDF&78.31$\pm$2.01&78.34$\pm$1.90&78.05$\pm$2.27&77.55$\pm$2.24&78.27$\pm$1.95\\
      FedAsync&80.75$\pm$1.51&80.43$\pm$1.24&81.00$\pm$1.55&79.95$\pm$1.36&80.61$\pm$1.49\\
      ours&\bf{84.20$\pm$2.39}&\bf{84.36$\pm$2.17}&\bf{84.40$\pm$1.93}&\bf{84.14$\pm$2.15}&\bf{83.94$\pm$2.26}\\
      \hline
    \end{tabular}}
\label{tab:comp}
\end{table*}

\begin{table*}[!t]
\small
\renewcommand{\arraystretch}{1.3}
  \centering
  \caption{Communication rounds and speedup ratio when reaching 70\% testing accuracy.}
  \setlength{\tabcolsep}{5mm}{
    \begin{tabular}{cccccc}
    \hline
      \multirow{2}{*}{\bf{Methods}}&\multicolumn{5}{c}{\bf{Stragglers (\%)}}\\
      \cline{2-6}
      &0&20&40&60&80\\
      \hline
      FedAvg&66&83&108&124&138\\
      FedProx&69 (0.96$\times$)&75 (1.11$\times$)&79 (1.37$\times$)&88 (1.41$\times$)&95 (1.45$\times$)\\
      FedDF&59 (1.12$\times$)&52 (1.60$\times$)&62 (1.74$\times$)&58 (2.14$\times$)&57 (2.42$\times$)\\
      FedAsync&83 (0.79$\times$)&84 (0.98$\times$)&83 (1.30$\times$)&87 (1.43$\times$)&82 (1.68$\times$)\\
      ours&\bf{39 (1.69$\times$)}&\bf{44 (1.89$\times$)}&\bf{44 (2.45$\times$)}&\bf{42 (2.95$\times$)}&\bf{40 (3.45$\times$)}\\
      \hline
    \end{tabular}}
\label{tab:comp2}
\end{table*}
\subsection{Comprehensive Performance Comparison}
To make a comprehensive comparison of the algorithm performance, several state-of-the-art frameworks that focus on different challenges of multi-party learning are introduced in the experiments, and these algorithms share similar ideas with the proposed framework. For addressing the statistical heterogeneity, FedProx \cite{Sahu2020FederatedOI} introduces a proximal term to the local subproblem to limit the impact of variable local updates, which allows for incorporating variable amounts of local work resulting from that. As an ensemble distillation for model fusion, FedDF \cite{Li2019FedMDHF} uses transfer learning and knowledge distillation to develop a framework, in which each party computes the class store on the public dataset and transmits the result to the central server. The number of public dataset is set the average number of local samples. FedAsync \cite{Sprague2018AsynchronousFL} is an asynchronous aggregation scheme for multi-party learning, where the server aggregates the model parameters immediately when receiving the client's parameters without waiting for any other clients. 

The experiments are conducted on a non-IID subset of the FEMNIST dataset, containing 100 clients and 18476 samples, and the number of rounds is 150. We employ the same model structure as MNIST and evaluate the performance in the system heterogeneity scenario by introducing random stragglers, whose updates will delay and a communication epoch means all clients uploaded their updates. All state-of-the-art approaches are available in the scene, while FedAvg merely drops these clients upon reaching the global clock cycle. It is a more realistic scenario that facilitates to measure the effectiveness and robustness of the methods. The testing accuracy is shown in Table.~\ref{tab:comp}, and the convergence speed comparison of algorithms is shown in Table.~\ref{tab:comp2}.

It is observed that the performance of FedAvg degrades significantly with the increasing number of stragglers, as fewer clients participate in multi-party learning in each communication round. The proximal term in FedProx is assistant to improve the stability by incorporate the partial updates from stragglers so that the performance decline more slowly compared with FedAvg. The asynchronous aggregation approaches, FedDF and FedAsync, are not affected by the number of stragglers and show stable performance. The proposed method is appliable to the asynchronous system as well and achieves superior performance with a various number of stragglers. Moreover, it converges faster than other algorithms and yields about twice the speedup ratio. FedDF also aggregates the outputs from clients instead of the model parameters, but it heavily relies on the public dataset, which is limited in quantity and quality. Thus, the convergence of FedAsync has a high rate but a low accuracy. 

\subsection{Effect of Incentive Mechanism}

\begin{figure}[!t]
  \centering
  \includegraphics[width=0.8\linewidth]{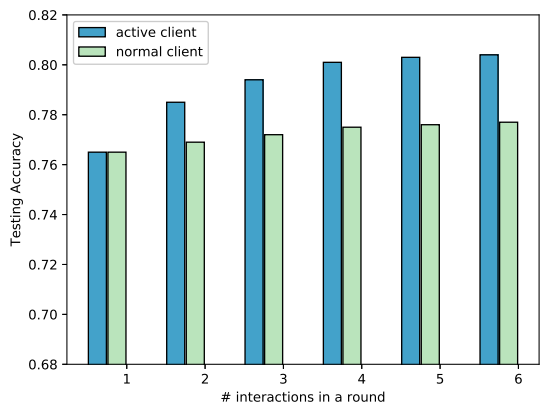}
  \caption{
  Testing accuracy of active client and normal client under different number of interactions on CIFAR-10.}
\label{fig:inc}
\end{figure}
\begin{table*}[tbp]
\small
\renewcommand{\arraystretch}{1.3}
  \centering
  \caption{Influence of different data quality levels on testing accuracy.}
  \setlength{\tabcolsep}{4mm}{
    \begin{tabular}{p{1mm}cccccccc}
    \cline{2-9}
      &\multirow{2}{*}{\bf{Methods}}&\multicolumn{7}{c}{\bf{Word Error Probability}}\\
      \cline{3-9}
      &&0&0.1&0.2&0.3&0.4&0.5&0.6\\
      \cline{2-9}
      \multirow{2}{*}{\rotatebox{90}{FedAvg}}&well-behaved&86.66&86.88&86.75&86.69&86.73&87.03&86.56\\
      &unreliable&86.74&86.24&85.77&85.19&84.68&83.99&83.13\\
        \cline{2-9}
        \multirow{3}{*}{\rotatebox{90}{ours}}&well-behaved&87.31&87.24&87.13&87.20&87.39&87.43&87.06\\
      &unreliable&87.11&86.52&86.01&85.37&84.20&82.74&80.43\\
      &honest&87.25&87.20&86.89&86.65&86.31&86.09&85.46\\
      \cline{2-9}
    \end{tabular}}
\label{tab:inq}
\end{table*}
\begin{table*}[tbp]
\small
\renewcommand{\arraystretch}{1.3}
  \centering
  \caption{Influence of malicious clients on testing accuracy.}
  \setlength{\tabcolsep}{4mm}{
    \begin{tabular}{p{1mm}cccccccc}
    \cline{2-9}
      &\multirow{2}{*}{\bf{Methods}}&\multicolumn{7}{c}{\bf{Attack Strength}}\\
      \cline{3-9}
      &&0&0.15&0.30&0.45&0.60&0.75&0.90\\
      \cline{2-9}
      \multirow{2}{*}{\rotatebox{90}{FedAvg}}&well-behaved&86.65&86.87&86.49&86.57&86.91&86.85&86.71\\
      &unreliable&86.76&86.08&85.61&84.14&82.30&80.89&79.07\\
        \cline{2-9}
        \multirow{3}{*}{\rotatebox{90}{ours}}&well-behaved&87.10&87.43&86.95&87.25&87.50&87.11&87.23\\
      &unreliable&87.29&81.58&71.26&58.65&46.19&29.43&10.91\\
      &honest&87.36&86.84&86.67&86.51&86.38&86.53&86.40\\
      \cline{2-9}
    \end{tabular}}
\label{tab:inc}
\end{table*}

The presented many-to-one knowledge sharing framework is naturally combined with the incentive mechanism, as the discriminative module deployed on the server has more affinity for clients that share more high-quality knowledge, while all clients in FedAvg are treated equally regardless of data quality or level of interaction, since all clients are with the same number of local samples. It is assumed that there exist 5 unreliable participants (with low-quality data or poisoned dataset) in 20 clients, and the experiments are conducted on the 20 Newsgroups dataset. For the FedAvg algorithm, the testing accuracy with unreliable clients participating in, denoted as 'unreliable', is compared with that obtained in a federation with all honest clients, named 'well-behaved', to measure the interference of these unreliable clients. Since the cognitive module is trained with local data and hence unique for each user in our scheme, 'unreliable' means testing accuracy on the 5 clients with low-quality data, while 'honest' represents that on the remaining 15 clients.

To quantify the data quality of local training data in unreliable clients, we introduce the word error probability to indicate the percentage of tokens in the corpus that are randomly replaced or deleted, and the experimental results are given in Table~\ref{tab:inq}. With the increase of the word error probability, all models experience varying degrees of performance degradation, and the value is positively correlated with the probability. Nevertheless, for unreliable clients that provide unstable models to the server, the testing accuracy of the FedAvg algorithm is better than that of ours, since the influence of migrated network parameters is diluted by the majority of normal parameters of uploaded by honest clients. As a contrast, the accuracy of these clients for the presented method reduces from 87\% to around 80\%, which means unreliable data providers suffer more performance penalty. On the contrary, the performance of honest clients in our scheme only drops by about 1\%, while the value in FedAvg is more than 3\%, indicating that we provide a more accountable knowledge aggregation strategy for all participants and encourage users to provide training data with higher quality.

For the malicious clients with poisoning attacks \cite{Bhagoji2019AnalyzingFL} \cite{Fang2020LocalMP}, some labels of the training samples are deliberately modified, i.e. the random label flipping attack, in order to mislead the global model. The attack strength represents the modified label percentage, and the experimental results are shown in Table~\ref{tab:inc}. Though exhibiting similar characteristics, the malicious clients are more destructive compared with low-quality data providers. For the FedAvg algorithm, the accuracy of the global model for all clients decreases by 7\% in the unreliable execution environment. In our scheme, the model performance of poisoning attackers and honest users are polarized, since that for the former approximates random guess when the attack strength is 0.9, while it is barely affected for the latter. Due to the misleading labels for the cognitive module, which is kept locally and is not corrected by normal parameters as FedAvg, the malicious clients cannot learn distinct features and take the consequences. The descriptive module cannot learn meaningful data distribution from the cognitive module, thus its shared cognition has a large trace of the covariance matrix $\Sigma$, and it is excluded from the knowledge sharing process on the server, and this is the reason for a higher accuracy for honest clients in a poisoning attack scenario than that when facing low-quality data providers. The mechanism makes the poisoning attacks futile to the majority of honest clients. Though it can effectively handle untarget poisoning attacks, it may not achieve the same performance when facing target attacks, which has a smaller variance, and the more effective solution is left in our future work. Besides, it provides participants with more autonomy in tagging to adapt their personalization in specific tasks, such as face recognition, so that the local model performs well on personalized data without affecting other users.

\begin{figure*}[!t]
  \centering
  \subfigure[MNIST]{
    \includegraphics[width=0.36\linewidth]{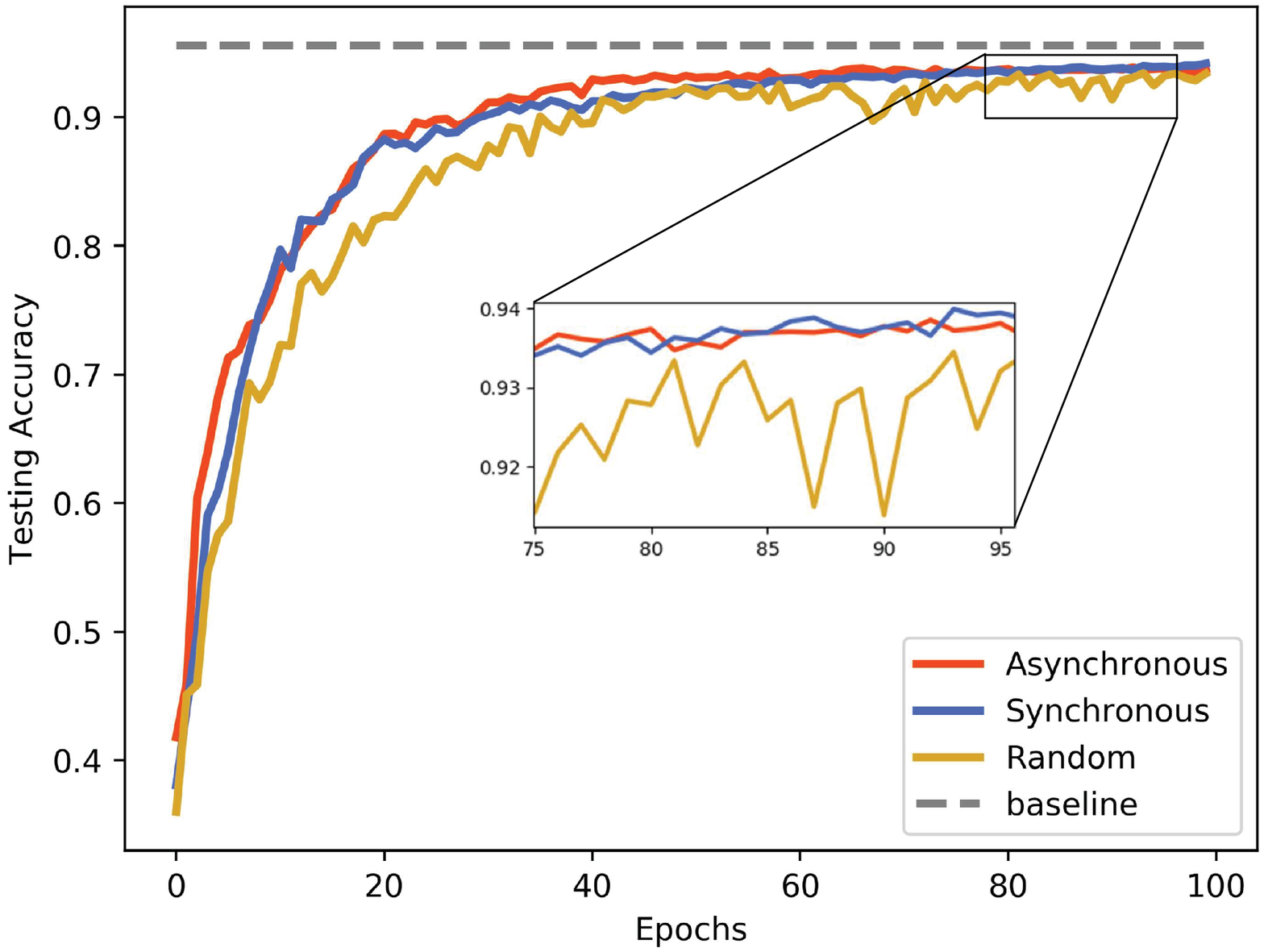}}
    \qquad\qquad
  \subfigure[SST-5]{
    \includegraphics[width=0.36\linewidth]{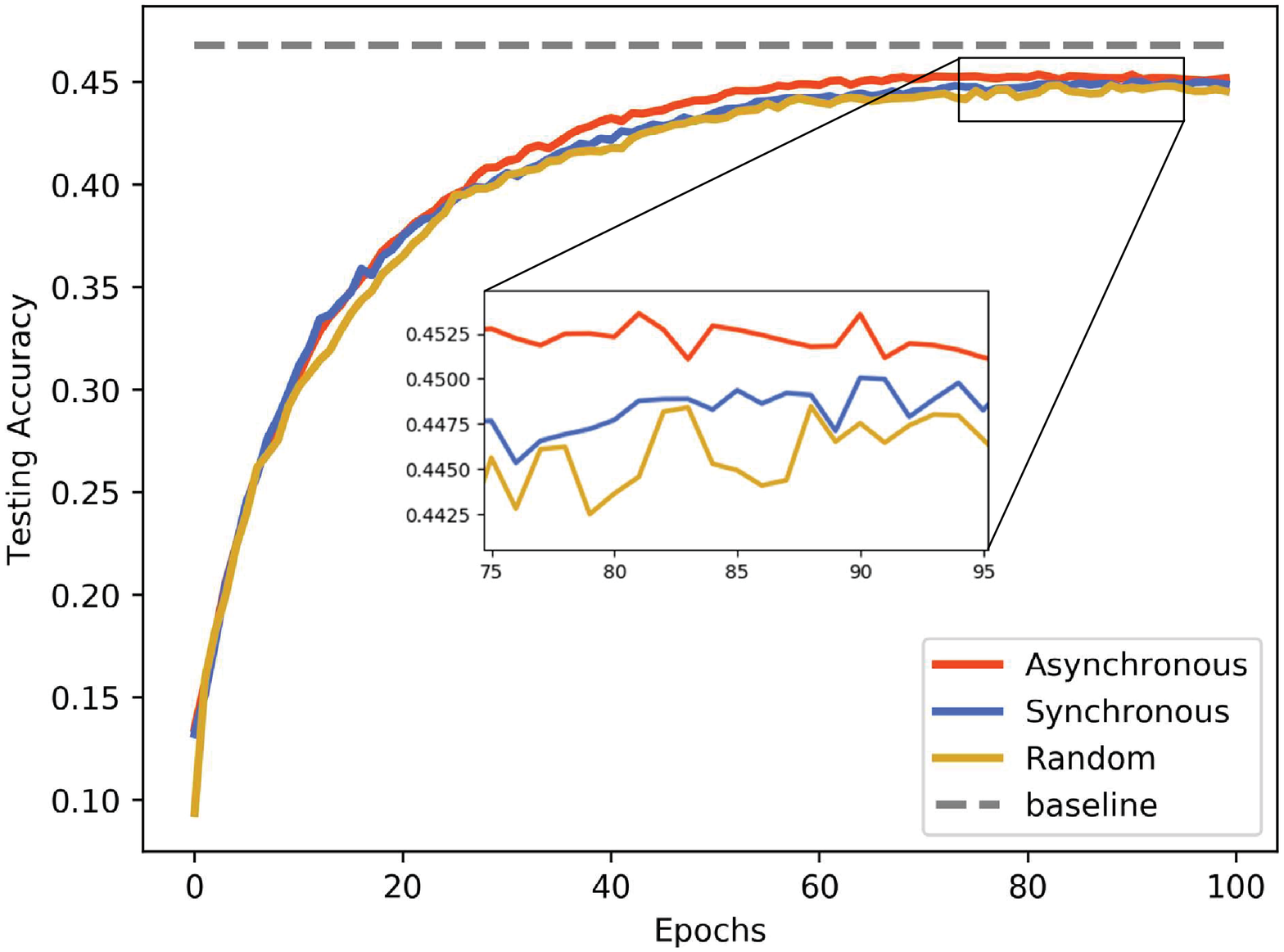}}
  \caption{Comparative studies of the proposed scheme in the asynchronous and synchronous settings on MNIST and SST-5.}
\label{fig:asy}
\end{figure*}
We also evaluate the effect of the number of interactions in a communication round on CIFAR-10, as shown in Fig.~\ref{fig:inc}. In asynchronous settings, we consider one round of communication to be all clients interacting at least once. Specifically, as our knowledge sharing scheme is not constrained by the synchronous parameter aggregation, we set 5 clients in 20 as active ones and interact with the server more frequently, while the others who interact once with the server in a round are normal clients. It is obvious that the accuracy of active clients increases with more interactions. As mentioned before, the central discriminative module has more connections with their imparted knowledge and has more affinity for active clients, and the local cognitive modules are trained more times to adapt the downloaded central discriminative module as well. Moreover, normal clients in the scenario perform slightly better when the number of interactions in a round is greater, because more training times on active models also accelerate the convergence of the central discriminative module.

\subsection{Performance in Asynchronous Setting}

We also conduct comparative studies of the proposed scheme in the asynchronous and synchronous settings on the MNIST and SST-5 dataset, as illustrated in Fig.~\ref{fig:asy}. Meanwhile, different from the asynchronous scenario where all clients upload and download in turn, we introduce a comparison with random interactions, and every 20 interactions are counted as one round of communication. 

It is observed that the proposed scheme demonstrates similar performance in the asynchronous environment to the synchronous setting. The asynchronous communication even achieves 0.5\% performance gains on the SST-5 dataset due to the updating policy of the representative collaborative cognition $\mathcal{R}$, which requires a minimum change interval that is hard for the synchronous communication to reach, because testing accuracy for all clients is close and may not meet the change criteria. For the same reason, the performance of the random method is slightly worse with poor robustness, especially on the MNIST dataset. 

\section{Conclusion}
\label{para:5}
In this work, we develop a novel contrastive knowledge sharing framework towards explainable multi-party learning. To address the problem of statistical and system heterogeneity in traditional multi-party setting, we introduce modularized functional zones for each client inspired by the process of human cognition and communication, which is capable of selectively imparting the acquired knowledge to the central server explicitly, and it prevents privacy disclosure of raw data from uploaded gradients. It achieves comparable performance to the conventional FedAvg algorithm in both IID and non-IID setting. Moreover, the separately trained modules provide natural choices for the incentive mechanism, and the scheme illustrates a more accountable knowledge aggregation strategy when facing unreliable clients and active clients. Extensive experimental results on several real-world datasets with various neural networks demonstrate its effectiveness in synchronous and asynchronous environments.

We will explore the following directions in the future:

(1) We provide a more explainable view for multi-party learning problems, which achieves comparable performance to conventional parameter aggregation approaches. Nevertheless, it is still a preliminary investigation in this direction, and the naive design of loss functions imposes restrictions on the performance of the framework at the current stage. Thus, we plan to carry out further research on the theoretical analysis and flexible module of the framework in the future.

(2) The proposed scheme can effectively handle a variety of scenarios, such as statistical heterogeneity and system heterogeneity. However, as the cognitive module is trained with local data thus unique for each user, the framework at the current stage may not well handle cold-start users with insufficient local samples \cite{Zhu2020AddressingTI}. We try to explore some new approaches for cognitive ability transferring, such as unsupervised encoding techniques \cite{Wang2015UnsupervisedJF}, to serve cold-start users without exposing private information.


%





\ifCLASSOPTIONcaptionsoff
  \newpage
\fi


\bibliographystyle{IEEEtran}
\bibliography{ref}
%




\end{document}